\documentclass[11pt]{article}
\usepackage{acl2015}
\usepackage{times}
\usepackage{url}
\usepackage{latexsym}

\usepackage{latexsym}
\usepackage{graphicx}
\usepackage{paralist}
\usepackage{amsmath}
\usepackage[round]{natbib}
\usepackage{url}
\usepackage{wrapfig}
\newcommand{\eg}{\emph{e.g.,}}

\newcommand{\ie}{\emph{i.e.,}}

\newcommand{\defoccur}[1]{\textsl{#1}}

\long\def\argmax{\arg\;\max}

\graphicspath{{images/}}
\DeclareGraphicsExtensions{.png,.jpg,.pdf}

\newif\iffinal
\finaltrue

\title{Robot Language Learning, Generation, and Comprehension}

\iffinal
\author{Daniel Paul Barrett \and
        Scott Alan Bronikowski \and
        Haonan Yu \and
        Jeffrey Mark Siskind\\
        Purdue University\\
        School of Electrical and Computer Engineering\\
	465 Northwestern Avenue\\
	West Lafayette, IN 47907-2035, USA\\
	\texttt{\{dpbarret,sbroniko,yu239,qobi\}@purdue.edu}}
\else
\author{anonymous authors}
\fi

\date{}

\begin{document}
\maketitle
\begin{abstract}
We present a unified framework which supports grounding natural-language
semantics in robotic driving.
This framework supports acquisition (learning grounded meanings of nouns and
prepositions from human annotation of robotic driving paths), generation (using
such acquired meanings to generate sentential description of new robotic
driving paths), and comprehension (using such acquired meanings to support
automated driving to accomplish navigational goals specified in natural
language).
We evaluate the performance of these three tasks by having independent human
judges rate the semantic fidelity of the sentences associated with paths,
achieving overall average correctness of 94.6\% and overall average
completeness of 85.6\%.
\end{abstract}

\section{Introduction}

\begin{figure*}
  \centering
  \resizebox{\textwidth}{!}{\begin{tabular}{@{}lc@{}}
  \begin{tabular}{@{}lccc@{}}
  \rotatebox{90}{\hspace*{-20pt}\textbf{acquisition}}&
  \begin{tabular}{@{}lc@{}}
    \rotatebox{90}{\hspace*{-10pt}\textbf{input}}&
    \resizebox{2\columnwidth}{!}{\begin{tabular}{@{}ccc@{}}
      \includegraphics[height=1in]{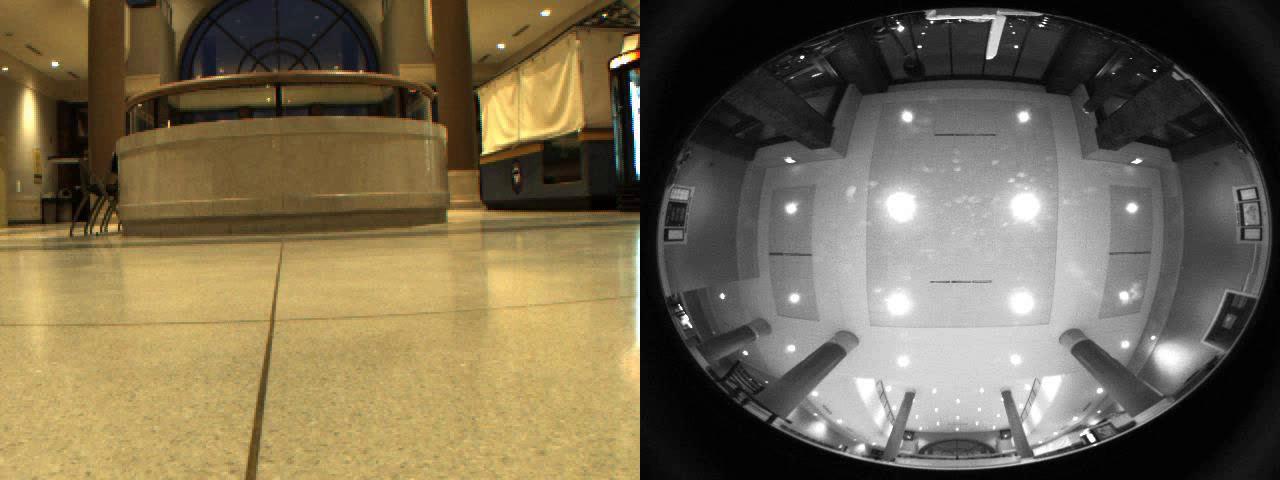}&
      \raisebox{20pt}{\begin{minipage}[b]{2in}
          \emph{The robot went in front of the bag which is left of the bag then
            went towards the chair.}
      \end{minipage}}&
      \includegraphics[height=1in]{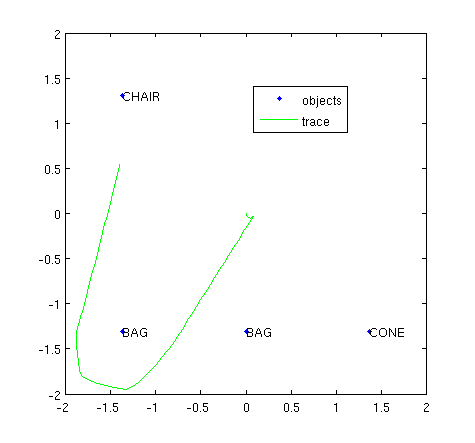}\\
    \end{tabular}}\\
    \rotatebox{90}{\hspace*{60pt}\textbf{output}}&
    \resizebox{2\columnwidth}{!}{\begin{tabular}{@{}rcccccc@{}}
        \rotatebox{90}{\makebox[0pt]{\hspace*{50pt}{\textbf{nouns}}}}&
        \includegraphics[width=0.15\textwidth]{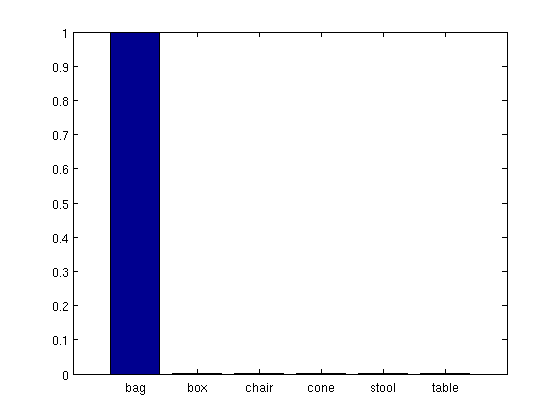}&
        \includegraphics[width=0.15\textwidth]{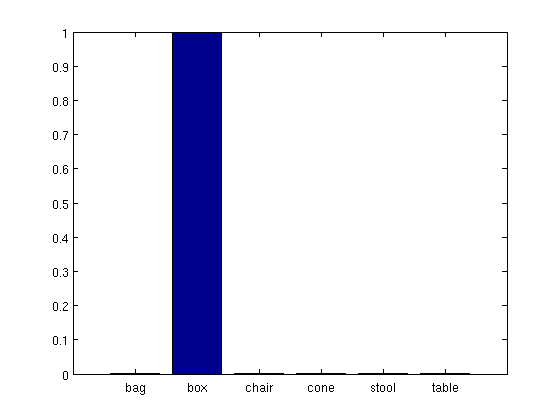}&
        \includegraphics[width=0.15\textwidth]{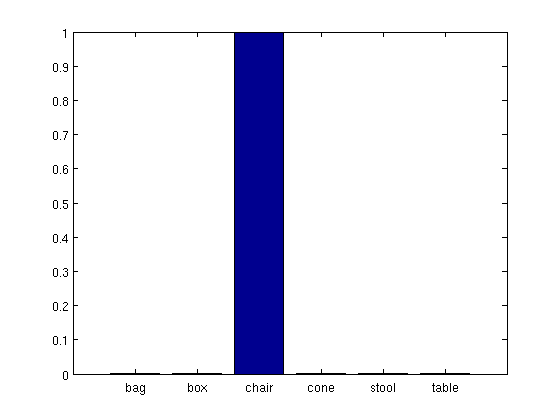}&
        \includegraphics[width=0.15\textwidth]{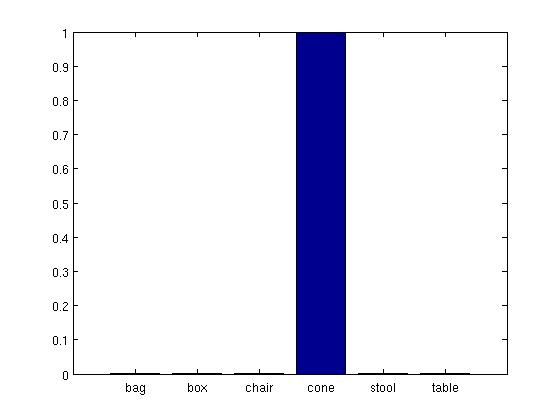}&
        \includegraphics[width=0.15\textwidth]{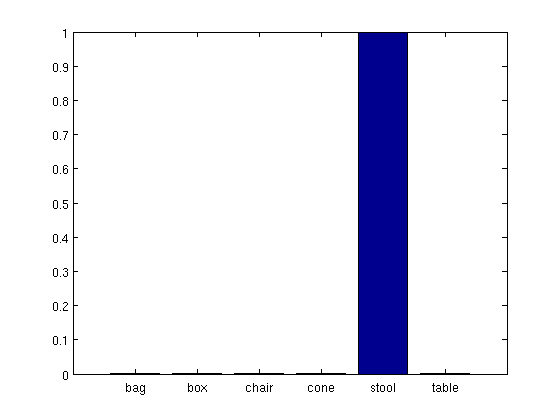}&
        \includegraphics[width=0.15\textwidth]{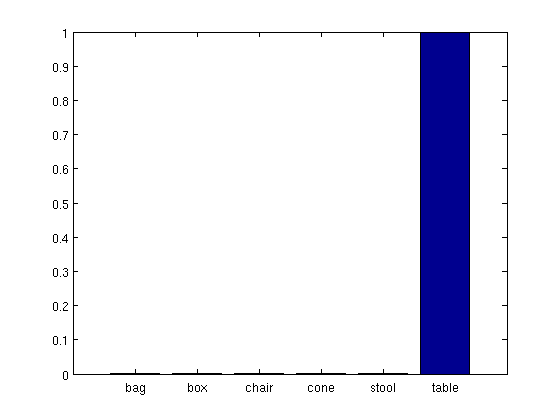}\\
        &
        \emph{bag} & \emph{box} & \emph{chair} & \emph{cone} & \emph{stool} &
        \emph{table}\\
        \rotatebox{90}{\makebox[0pt]{\hspace*{65pt}{\textbf{position}}}}&
        \includegraphics[height=1in]{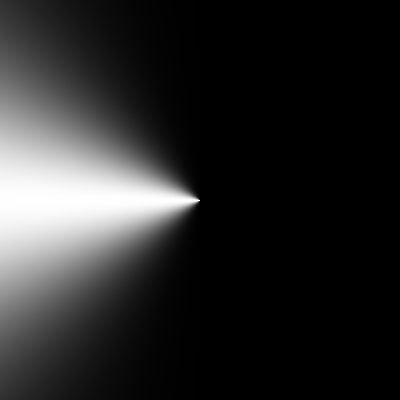}&
        \includegraphics[height=1in]{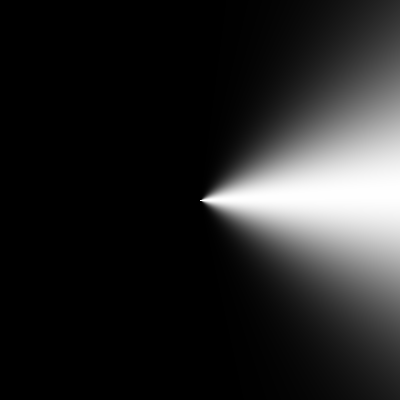}&
        \includegraphics[height=1in]{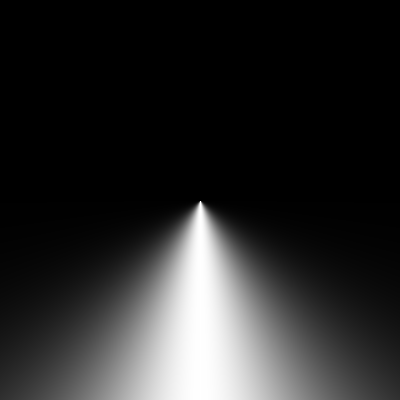}&
        \includegraphics[height=1in]{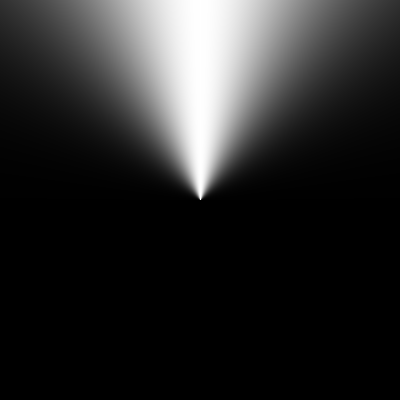}&
        \includegraphics[height=1in]{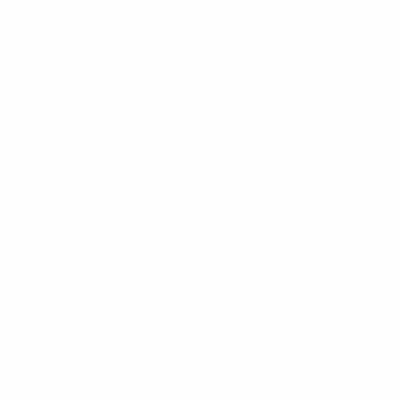}&
        \includegraphics[height=1in]{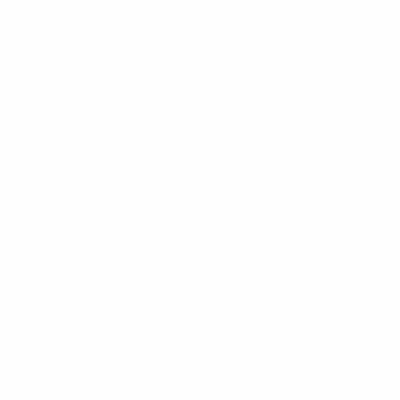}\\
        \rotatebox{90}{\makebox[0pt]{\hspace*{65pt}{\textbf{velocity}}}}&
        \includegraphics[height=1in]{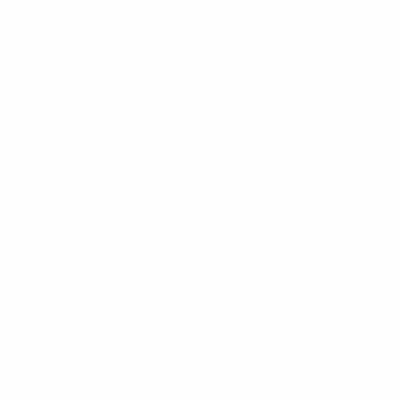}&
        \includegraphics[height=1in]{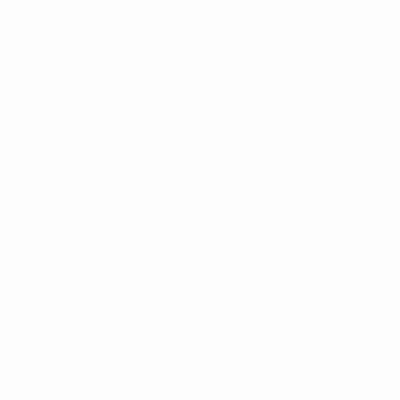}&
        \includegraphics[height=1in]{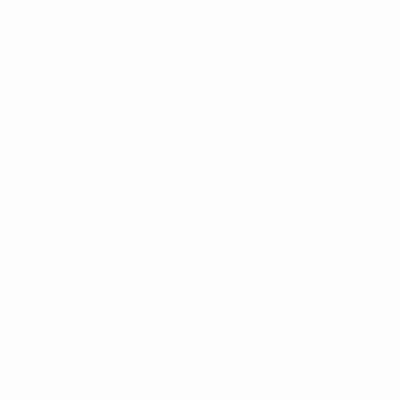}&
        \includegraphics[height=1in]{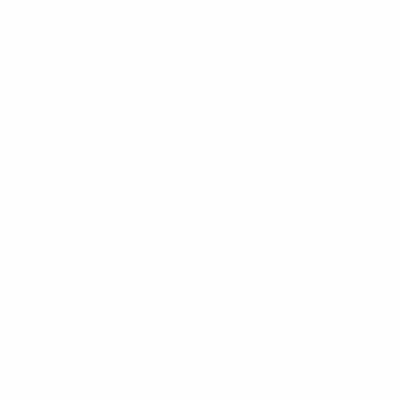}&
        \includegraphics[height=1in]{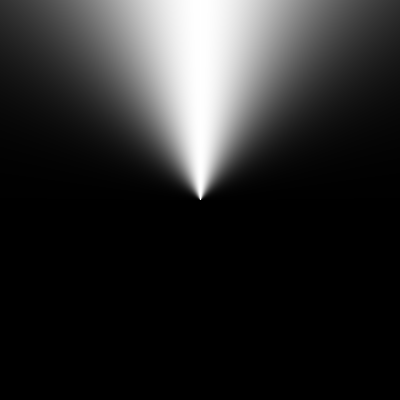}&
        \includegraphics[height=1in]{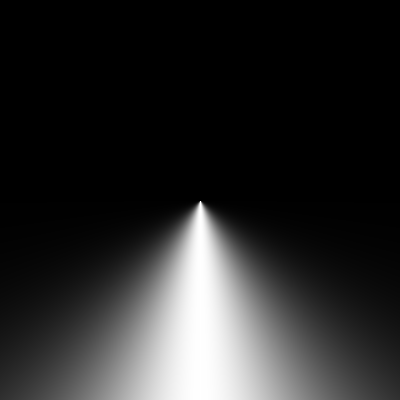}\\
        &
        \emph{left of}&
        \emph{right of}&
        \emph{in front of}&
        \emph{behind}&
        \emph{towards}&
        \emph{away from}
    \end{tabular}}
  \end{tabular}&
  $\Longrightarrow$&
  \begin{tabular}{@{}lc@{}}
  \rotatebox{90}{\hspace*{-20pt}\textbf{generation}}&
  \begin{tabular}{@{}lc@{}}
    \rotatebox{90}{\hspace*{-10pt}\textbf{input}}&
    \resizebox{\columnwidth}{!}{\begin{tabular}{@{}cc@{}}
        \includegraphics[height=1in]{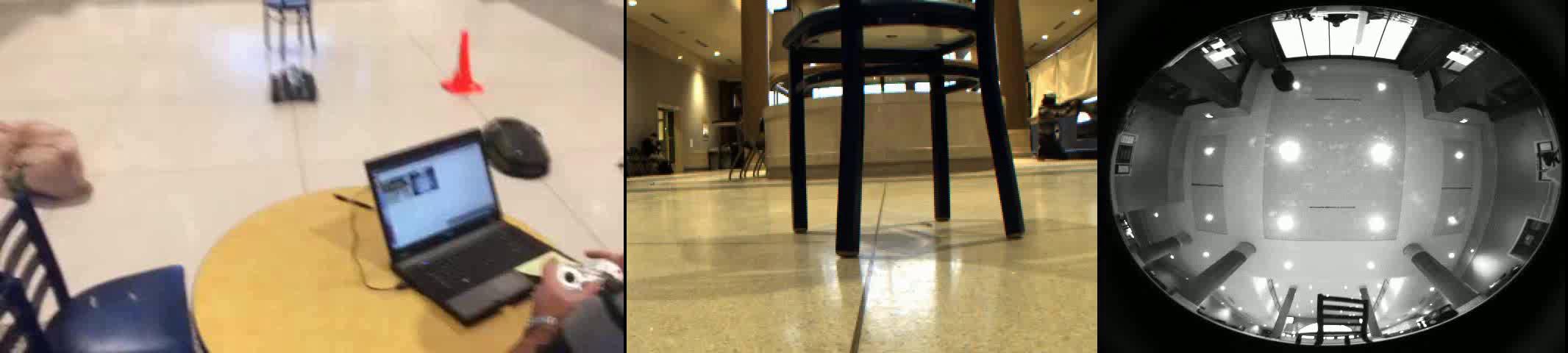}&
        \includegraphics[height=1in]{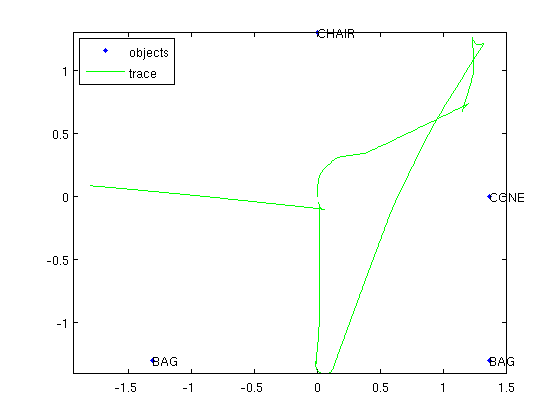}
    \end{tabular}}\\
    \\
    \rotatebox{90}{\hspace*{-15pt}\textbf{output}}&
    \begin{minipage}{\columnwidth}
      \emph{The robot went in front of the chair then went
        away from the chair and
        behind the cone then went right of the bag which is left of the cone
        then went left of the bag which is in front of the cone then went
        away from the cone and away from the chair.}
      \end{minipage}
  \end{tabular}\\
  $+$&\raisebox{2.5pt}{\rule{3.5in}{1pt}}\\
  \rotatebox{90}{\hspace*{-30pt}\textbf{comprehension}}&
  \begin{tabular}{@{}lc@{}}
    \rotatebox{90}{\hspace*{-10pt}\textbf{input}}&
    \begin{minipage}{\columnwidth}
      \emph{The robot went behind the bag which is in front of the
            bag then went in front of the bag which is left of the chair then
            went towards the cone then went away from the chair then went right
            of the chair then went right of the bag which is left of the cone.}
      \end{minipage}\\
    \\
    \rotatebox{90}{\hspace*{-10pt}\textbf{output}}&
    \resizebox{\columnwidth}{!}{\begin{tabular}{@{}cc@{}}
        \includegraphics[height=1in]{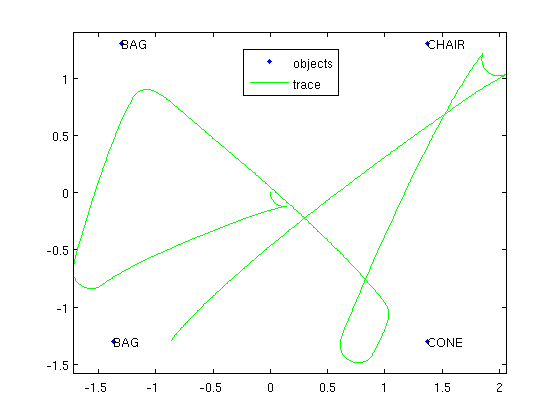}&
      \includegraphics[height=1in]{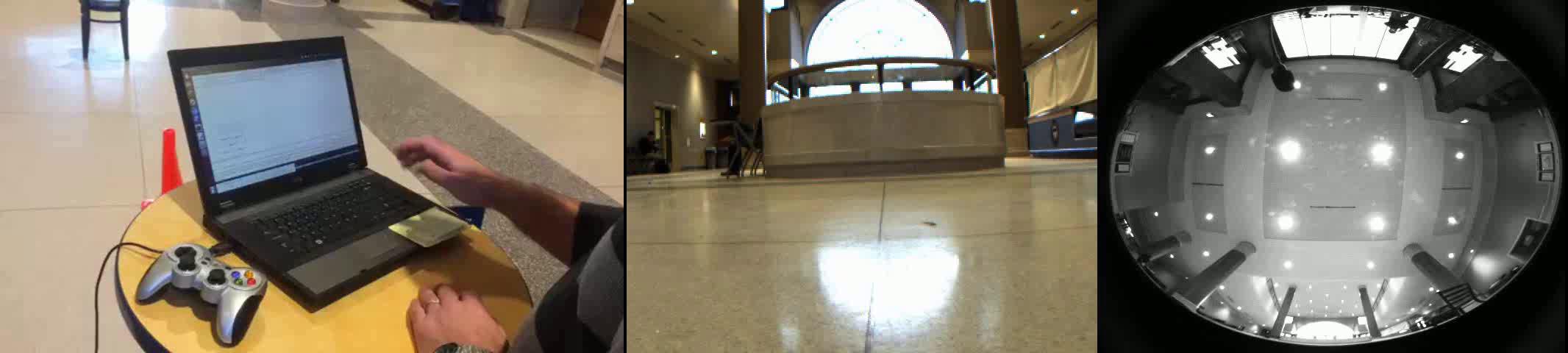}
      \end{tabular}}
  \end{tabular}
  \end{tabular}
  \end{tabular}
  \end{tabular}}
  \par\vspace*{-1ex}
  \caption{(left)~A human drives the mobile robot through paths according to
    sentential instructions while odometry reconstructs the robot's paths. 
    This allows the robot to learn the meanings of the nouns and prepositions.
    Hand-designed models are shown here for reference; actual learned
    models are shown in Fig.~\ref{fig:experiments}.
    Note that the distributions are uniform in velocity angle (bottom row) for
    \emph{left of}, \emph{right of}, \emph{in front of}, and \emph{behind} and
    in position angle (top row) for \emph{towards} and \emph{away from}.
    These learned meanings support generation of English descriptions of
    new paths driven by teleoperation (top right) and autonomous driving of
    paths that meet navigational goal specified in English descriptions
    (bottom right).} 
  \vspace*{-2ex}
  \label{fig:overview}
\end{figure*}

With recent advances in machine perception and robotic automation, it becomes
increasingly relevant and important to allow machines to interact with humans
in natural language in a \emph{grounded fashion}, where the language refers to
actual things and activities in the world.
Here, we present our efforts to automatically drive---and learn to drive---a
mobile robot under natural-language command.
Our contribution is summarized in Fig.~\ref{fig:overview}.
A human teleoperator is given a set of sentential instructions designating
robot paths.
The operator then drives a mobile robot under radio control according to these
instructions through a variety of floorplans.
The robot uses onboard odometry and inertial guidance sensors to determine its
location in real time and saves traces of the driving path to log files.
From a training corpus of paths paired with sentential descriptions and
floorplan specifications, our system automatically learns the meanings of
nouns that refer to objects in the floorplan and prepositions that describe
both the spatial relations between floorplan objects and between such objects
and the robot path.
With such learned meanings, the robot can then generate sentential descriptions
of new driving activity undertaken by the teleoperator.
Moreover, instead of manually controlling the robot through teleoperation, one
can issue the robot natural-language commands which can induce fully automatic
driving to satisfy the path specified in the natural-language command.

We have conducted experiments with an actual radio-controlled robot that
demonstrate all three of these modes of operation: acquisition, generation, and
comprehension.
We demonstrate successful completion of all three of these tasks on hundreds
of driving examples.
We evaluate the fidelity of the sentential descriptions produced automatically
in response to manual driving and the fidelity of the driving paths induced
automatically to fulfill natural-language commands, by presenting the pairs of
sentences together with the associated paths to human judges.
Overall, the average ``correctness'' (the degree to which the description is
true of the path) reported is 94.6\% and the average ``completeness'' (the
degree to which the description fully covers the path) reported is 85.6\%.

\section{Related Work}
We know of no other work which presents a physical robot which learns word
meanings from physical robot paths paired with sentences, uses these learned
meanings to generate sentential descriptions of manually driven paths, and
automatically plans and physically drives paths to satisfy input sentential
descriptions. 

While there is other work which claims to learn the meanings of words from robot
paths or follow natural instructions, upon further inspection these systems
operate only within discrete simulation, as they utilize the internal
representation of the simulation to obtain discrete symbolic
primitives \citep{Tellex2014, Tellex2011, Kollar2010, Chen2011,Macmahon2006,Koller2010}.
Their space of possible robot actions, positions and states are very small and
are represented in terms of symbolic primitives like 
\textsc{turn left}, \textsc{turn right}, and \textsc{move forward $N$ steps}
\citep{Chen2011}, or \textsc{drive to location 1} and \textsc{pick up pallet 1}
\citep{Tellex2014}. 
Thus, they take a sequence of primitives like \{\textsc{drive to location 1};
\textsc{pick up pallet 1}\} and a sentence like \emph{go to the pallet and
  pick it up} and learn that the word \emph{pallet} maps to the primitive
\textsc{pallet}, that the phrase \emph{pick up} maps to the primitive
\textsc{pick up}, and that the phrase \emph{go to X} means \textsc{drive to
  location X}. 
In contrast, our robot and environment, being in the continuous physical
world, can take an uncountably infinite number of configurations.
We take a set of sentences matched with paths of the robot as input, where the
paths are densely sampled points in the real 2D Cartesian plane.
Not all points in the path correspond to words in the sentences, multiple
(often undescribed) relationships can be true of any point, and the
correspondence between described relationships and path points is unknown. 
This is a vastly more difficult problem.

Furthermore, previous work does not even solve the simplified problem without
additional annotation.
\citet{Kollar2010} requires hand-drawn positive and negative paths depicting
specific word meanings.
\citet{Tellex2011} requires manual annotation of the groundings of all words in
the training sentences to specific objects and relationships in the training
data. 
\citet{Tellex2014} does not require annotation of the grounding of each word,
but does require manual temporal segmentation and alignment of paths and the
pieces of multi-part sentences, whereas our method can learn without any such annotation.
%

\citet{Dobnik2005} has an actual robot but only learns to classify simple
phrases like \emph{A is near B} from robot paths paired with such phrases that
have hand-grounded nouns.
They can neither generate sentences describing driven paths, nor automatically
drive a path described by a sentence. 
Our system can do both of these, as well as learn meanings for both nouns and
prepositions.

\section{Our Mobile Robot}

\begin{figure}
  \centering
  \includegraphics[width=\columnwidth]{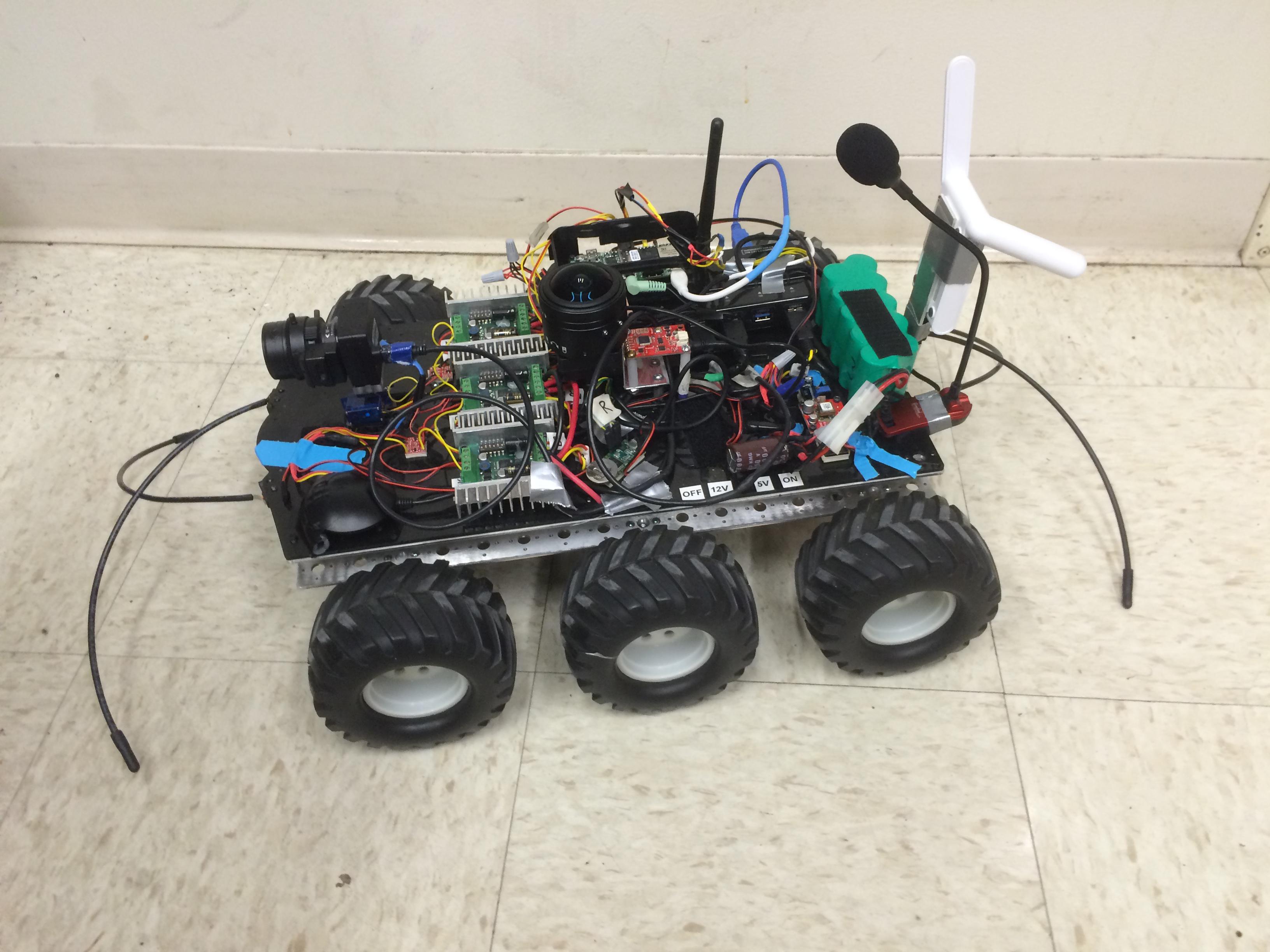}
  \par\vspace*{-2ex}
  \caption{Our custom mobile robot.}
  %
  %
  \label{fig:robot}
\end{figure}

All experiments were performed on a custom mobile robot
(Fig.~\ref{fig:robot}). 
This robot can be driven by a human teleoperator or drive itself automatically
to accomplish specified navigational goals. 
During all operation, robot localization is performed onboard the robot in
real-time via an Extended Kalman Filter \citep{Jazwinski1970} with odometry
from shaft encoders on the wheels and inertial-guidance from an IMU.

Due to sensor noise and mechanical factors such as wheel sliding, this
localization is noisy, but generally within 20cm of the actual location.
The video feed, localization, and all sensor and actuator data is logged in a
time-stamped format.
When conducting experiments on generation and acquisition, a human
teleoperator drives the robot along a variety of paths in a variety of
floorplans.
The path recovered from localization supports generation and acquisition.
When conducting experiments on comprehension, the path is first planned
automatically, then the robot automatically follows its planned path by
comparing the new odometry gathered in real time with the planned path and
controlling the wheels accordingly.

The use of an actual robot with noisy real-world sensor data increases the
difficulty of the tasks when compared to work which occurs in simulation.
The noisy robot position is densely sampled in the continuous domain.
For acquisition and generation, this adds an additional layer of uncertainty,
as the correspondence between individual points in the robot path and the
phrases of a sentence is unknown.

\section{Technical Details}

\subsection{Grammar and Logical Form}

We employ the grammar shown in Fig.~\ref{fig:grammar}, which, while small,
supports an infinite set of possible utterances, unlike the grammars used in
\citet{Teller2010} and \citet{Harris2005}.
Nothing turns on this however.
In principle, one could replace this grammar with any other mechanism for
generating logical form.
%
%
This paper concerns itself with semantics, not syntax, and only addresses
issues relating to the grounding of logical form.
This particular grammar is simply a convenient surface representation of our
logical form.

\begin{figure}
\resizebox{.7\columnwidth}{!}{
  \parbox{\columnwidth}{
  \begin{math}\displaystyle
    \begin{array}{l@{\hspace{2pt}}@{\hspace{2pt}}cl}
      \textrm{S}&\rightarrow&\textit{The robot}\;\textrm{VP}\\
      \textrm{VP}&\rightarrow&
      \textit{went}\;\textrm{PP}_{\textrm{path}}\;[\textit{then}\;\textrm{VP}]\\
      \textrm{PP}_{\textrm{path}}&\rightarrow&
      \textrm{P}_{\textrm{path}}\;\textrm{NP}\;
             [\textit{and}\;\textrm{PP}_{\textrm{path}}]\\
      \textrm{NP}&\rightarrow&
      \textit{the}\;\textrm{N}\;[\textrm{PP}_{\textrm{SR}}]\\
      \textrm{PP}_{\textrm{SR}}&\rightarrow&
      \textit{which is}\;\textrm{P}_{\textrm{SR}}\;\textrm{NP}\;
             [\textit{and}\;\textrm{PP}_{\textrm{SR}}]\\
      \textrm{P}_{\textrm{path}}&\rightarrow&
      \textit{left of}\mid\textit{right of}\mid
      \textit{in front of} \mid\textit{behind} \mid\textit{towards}\mid\textit{away from}\\
      \textrm{P}_{\textrm{SR}}&\rightarrow&
      \textit{left of}\mid\textit{right of}
      \mid\textit{in front of}\mid\textit{behind}\\
      \textrm{N}&\rightarrow&\textit{bag}\mid\textit{box}
      \mid\textit{chair}\mid\textit{cone}\mid\textit{stool}\mid\textit{table}
    \end{array}
  \end{math}
}
}
\par\vspace*{-2ex}
  \caption{The grammar used by our implementation.}
  \vspace*{-2ex}
  \label{fig:grammar}
\end{figure}

Note that our surface syntax allows two uses of prepositions (and the
associated prepositional phrases): as modifiers to nouns in noun phrases,
indicated with a subscript `SR' (\ie\ spatial relation), and as adjuncts to
verbs in verb phrases, indicated with a subscript `path.'
Many prepositions can be used in both SR and path form.
They share the same semantic representation and both uses are learned from the
pooled data of both kinds of occurrences in the training corpus.
Furthermore, note that the grammar supports infinite NP recursion: noun phrases
can contain prepositional phrases that, in turn, contain noun phrases.
Finally, note that the grammar supports conjunctions of prepositional phrases
in both SR and path form.

\begin{figure}[t]
  \centering
  \resizebox{\columnwidth}{!}{\begin{tabular}{@{}cc@{}}
    \includegraphics[height=1in]{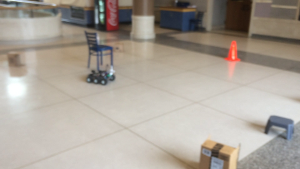}&
    \includegraphics[height=1in]{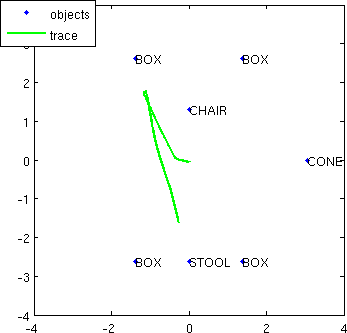}
  \end{tabular}}
  \par\vspace*{-2ex}
  \caption{Sample floorplan with robot path. (left)~Extrinsic image taken
    during operation. 
    (right)~Internal representation of floorplan consisting of labeled
    waypoints and localized path consisting of unlabeled waypoints.}
  \vspace*{-2ex}
  \label{fig:floorplan}
\end{figure}

We employ the logical form shown in Fig.~\ref{fig:logical-form}.
Informally, formulas in logical form denote paths through a floorplan.
Both paths and floorplans are specified as collections of waypoints.
A \defoccur{waypoint} is a 2D Cartesian coordinate optionally labeled with the
class of the object that resides at that coordinate,
\eg\ $(3,47,\textbf{bag})$
The waypoint is unlabeled, \eg $(3,47)$, if no object resides at that
coordinate.
A \defoccur{floorplan} is a set of labeled waypoints, while a \defoccur{path}
is a sequence of unlabeled waypoints (Fig.~\ref{fig:floorplan} right).
A formula in logical form contains three parts: a \defoccur{path
  quantifier}, a \defoccur{floorplan quantifier}, and a \defoccur{condition}
that the path through the floorplan must satisfy.
The condition is a conjunction of atomic formulas, predicates applied to
variables bound by the path or floorplan quantifiers.
The formula must be closed, \ie\ every variable in the condition must appear
either in the path quantifier or the floorplan quantifier.
The model of a formula is a set of bindings for each of the quantified path
variables to unlabeled waypoints, and floorplan variables to labeled waypoints.

\begin{figure}
  \centering
  \resizebox{.75\columnwidth}{!}
  {\begin{math}
    \begin{array}{lcl}
      \langle\emph{formula}\rangle&\rightarrow&
      \langle\emph{path quantifier}\rangle
      \langle\emph{floorplan quantifier}\rangle\\
      &&\langle\emph{atomic formula}\rangle
      (\wedge\langle\emph{atomic formula}\rangle)^*\\

      \langle\emph{path quantifier}\rangle&\rightarrow&
      [\langle\emph{var}\rangle(;\langle\emph{var}\rangle)^*]\\
      \langle\emph{floorplan quantifier}\rangle&\rightarrow&
      \{\langle\emph{var}\rangle(,\langle\emph{var}\rangle)^*\}\\
      \langle\emph{atomic formula}\rangle&\rightarrow&
      \langle\emph{atomic formula}_1\rangle\\
      &\mid&\langle\emph{atomic formula}_2\rangle\\
      \langle\emph{atomic formula}_1\rangle&\rightarrow&
      \textsc{bag}(\langle\emph{var}\rangle)\\
      &\mid&\textsc{box}(\langle\emph{var}\rangle)\\
      &\mid&\textsc{chair}(\langle\emph{var}\rangle)\\
      &\mid&\textsc{cone}(\langle\emph{var}\rangle)\\
      &\mid&\textsc{stool}(\langle\emph{var}\rangle)\\
      &\mid&\textsc{table}(\langle\emph{var}\rangle)\\
      \langle\emph{atomic formula}_2\rangle&\rightarrow&
      \textsc{leftOf}(\langle\emph{var}\rangle,
                      \langle\emph{var}\rangle)\\
      &\mid&\textsc{rightOf}(\langle\emph{var}\rangle,
                             \langle\emph{var}\rangle)\\
      &\mid&\textsc{inFrontOf}(\langle\emph{var}\rangle,
                               \langle\emph{var}\rangle)\\
      &\mid&\textsc{behind}(\langle\emph{var}\rangle,
                            \langle\emph{var}\rangle)\\
      &\mid&\textsc{towards}(\langle\emph{var}\rangle,
                             \langle\emph{var}\rangle)\\
      &\mid&\textsc{awayFrom}(\langle\emph{var}\rangle,
                              \langle\emph{var}\rangle)\\
    \end{array}
  \end{math}}
  \par\vspace*{-2ex}
  \caption{The logical form used by our implementation.}
  \label{fig:logical-form}
\end{figure}

The one-argument atomic formulas constrain the class of waypoints to which the
variables that appear as their arguments are bound.
The two-argument atomic formulas constrain the spatial relations between pairs
of waypoints to which the variables that appear as their arguments are bound.
The logical form in Fig.~\ref{fig:logical-form} contains a particular set of
six one-argument predicate and six two-argument predicates.
Nothing turns on this however.
This is simply the set of predicates that we use in the experiments reported.
The framework clearly extends to any number of predicates of any arity,
particularly since we learn the meanings of the predicates.

Straightforward (semantic) parsing and surface generation techniques map
bidirectionally between the surface language form as specified by the grammar
in Fig.~\ref{fig:grammar} and the logical form in Fig.~\ref{fig:logical-form}.
For example, a surface form like
\begin{quote}
\small{\emph{The robot went towards the stool, then went behind the chair
  which is right of the stool, then went towards the cone, then went away from
  the chair which is left of the cone, then went in front of the table.}}
\end{quote}
(commas added for legibility) would correspond to the following logical form:
\begin{equation}
\label{eq:logicalform}
\resizebox{.8\columnwidth}{!}
{\begin{math}
    [\alpha,\beta,\gamma,\delta,\epsilon]
    \{t,u,v,w,x,y,z\}
    \left(\begin{array}{l}
      \textsc{towards}(\alpha,t)\wedge
      \textsc{stool}(t)\wedge\\
      \textsc{behind}(\beta,u)\wedge
      \textsc{chair}(u)\wedge
      \textsc{rightOf}(u,v)\wedge
      \textsc{stool}(v)\wedge\\
      \textsc{towards}(\gamma,w)\wedge
      \textsc{cone}(w)\wedge\\
      \textsc{awayFrom}(\delta,x)\wedge
      \textsc{chair}(x)\wedge
      \textsc{leftOf}(x,y)\wedge
      \textsc{cone}(y)\wedge\\
      \textsc{inFrontOf}(\epsilon,z)\wedge
      \textsc{table}(z)
    \end{array}\right)
\end{math}}
\end{equation}
Note that in the above, nouns all correspond to one-argument predicates while
prepositions all correspond to two-argument predicates. 
But nothing turns on this.
One could imagine lexical prepositional phrases, like \emph{leftward}, that
correspond to one-argument predicates.
Moreover, path uses of prepositions specify waypoints in the path.
These appear in logical form as predicates whose first argument is a variable
in the path quantifier.
Similarly, SR uses of prepositions specify waypoints in the floorplan.
These appear in logical form as predicates whose first argument is a variable
in the floorplan quantifier.
Thus, in the above, the atomic formulas $\textsc{towards}(\alpha,t)$,
$\textsc{behind}(\beta,u)$, $\textsc{towards}(\gamma,w)$,
$\textsc{awayFrom}(\delta,x)$, and $\textsc{inFrontOf}(\epsilon,z)$ constitute
path uses while the atomic formulas $\textsc{rightOf}(u,v)$ and
$\textsc{leftOf}(x,y)$ constitute SR uses.
Note that each (path) prepositional phrase consists of a subset of the
atomic formulas in the condition, as indicated above by the line breaks.

\subsection{Representation of the Lexicon}
\label{sec:lexicon}
The lexicon specifies the meanings of the one- and two-argument predicates in
logical form.
The meanings of one-argument predicates are discrete distributions over the set
of class labels.
Note that the one-argument predicates, like \textsc{bag}, are distinct from the
class labels, like \textbf{bag}.
The mapping between such is learned.
Moreover, a given floorplan might have multiple instances of objects of the
same class.
These would be disambiguated with complex noun phrases such as
\emph{the chair which is right of the stool} and \emph{the chair which is left
  of the cone}.
Such disambiguating prepositional phrase modifiers of noun phrases can be
nested and conjoined arbitrarily.
Similarly, waypoints can be disambiguated by conjunctions of prepositional
phrase adjuncts.

Two-argument predicates specify relations between target objects and
reference objects.
In SR uses, the reference object is the object of the preposition while the
target object is the head noun.
For example, in \emph{the chair to the left of the table}, \emph{chair} is
the target object and \emph{table} is the reference object.
In path uses, the target object is a waypoint in the robot path while the
reference object is the object of the preposition.
For example, in \emph{went towards the table}, \emph{table} is the reference
object.
The lexical entry for each two-argument predicate is specified as the
location~$\mu$ and concentration~$\kappa$ parameters for multiple independent
von~Mises distributions \citep{Abramowitz1972} for a variety of angles between
target and reference objects.

\begin{figure}
  \centering
  \resizebox{\columnwidth}{!}{\begin{tabular}{@{}cc@{}}
    \includegraphics[width=0.35\columnwidth]{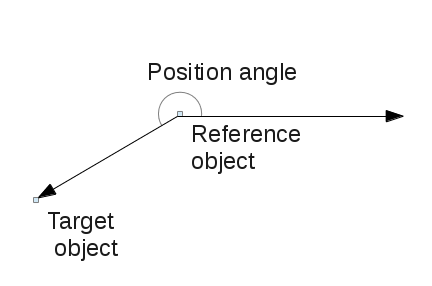}&
    \includegraphics[width=0.35\columnwidth]{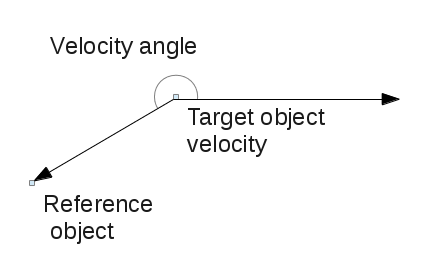}
  \end{tabular}}
  \par\vspace*{-2ex}
  \caption{(left)~How position angles are measured.
    (right)~How velocity angles are measured.}
  \vspace*{-2ex}
  \label{fig:angle}
\end{figure}

The meanings of two-argument predicates are specified as a pair of von~Mises
distributions on angles.
One, the \defoccur{position angle}, is the orientation of a vector from the
coordinates of the reference object to the coordinates of the target object
(Fig.~\ref{fig:angle} left).\footnote{Without loss
of generality, angles are measured in the frame of reference of the robot
prior to the beginning of action, which is taken to be the origin.}
%
The same distribution is used both for SR and path uses.
The second, the \defoccur{velocity angle}, is the angle between the velocity
vector at a waypoint and a vector from the coordinates of the waypoint to the
coordinates of the reference object (Fig.~\ref{fig:angle} right).
This is only used for path uses, because it requires computation of the
direction of robot motion which is determined from adjacent waypoints in the
path.
This angle is thus taken from the frame of reference of the robot.

Fig.~\ref{fig:overview}(bottom left) illustrates how this framework is used to
represent the meanings of prepositions.
Here, we render the angular distributions as potential fields around the
reference object at the center for the position angle, and the target object at
the center for the velocity angle.
The intensity of a point (target object for position angle) reflects its
probability mass.
Note that the distributions are uniform in velocity angle for \emph{left of},
\emph{right of}, \emph{in front of}, and \emph{behind} and in position angle
for \emph{towards} and \emph{away from}.

\subsection{Tasks}

We formulate sentential semantics as a variety of relationships between a
sentence~$\mathbf{s}$, or more precisely a formula in logical form, a
path~$\mathbf{p}$, a sequence of unlabeled waypoints, a floorplan~$\mathbf{f}$,
a set of labeled waypoints, and a lexicon~$\Lambda$, the collective~$\mu$
and~$\kappa$ parameters for the angular distributions for each of the
two-argument predicates and the discrete distributions for each of the
one-argument predicates.
\begin{compactdesc}
\item[acquisition] Learn a lexicon~$\Lambda$ from a collection of
  observed paths~$\mathbf{p}_i$ taken by the robot in the corresponding
  floorplans~$\mathbf{f}_i$ as described by human-generated
  sentences~$\mathbf{s}_i$. 
\item[generation] Generate a sentence~$\mathbf{s}$ that describes an
  observed path~$\mathbf{p}$ taken by the robot in a given
  floorplan~$\mathbf{f}$ with a known lexicon~$\Lambda$.
\item[comprehension] Generate a path~$\mathbf{p}$ to be taken by the
  robot that satisfies a given sentence~$\mathbf{s}$ issued as a command in a
  given floorplan~$\mathbf{f}$ with a known lexicon~$\Lambda$.
\end{compactdesc}

\subsubsection{Acquisition}

To perform acquisition, we formulate a large hidden Markov model (HMM), with a
state $k$ for every path prepositional phrase $\textrm{PP}_{\textrm{path},k}$ in
each sentence in the training corpus.
The observations for this HMM are the sequences of path waypoints in the
training corpus.
Each state's output model sums over all mappings $m$ between object references in the
$\textrm{PP}_{\textrm{path},k}$ and floorplan waypoints.
Given such a mapping, the output model for a state $k$ consists of the product of
the probabilities $P$ determined by each atomic formula $i$ in the logical form derived
from~$\textrm{PP}_{\textrm{path},k}$, given the probability models for the
predicates as specified by the current estimates of the parameters in~$\Lambda$: 
  \begin{equation}
    \vspace*{-1ex}
    \begin{array}{l}
    R_k(\mathbf{PP_{path,k}},\mathbf{p},\mathbf{f},\Lambda,m) = \\
    \hspace{2ex}\prod_{i} P(w_{a_{i0}}\ldots w_{a_{iN_i}} | \Lambda_i, m)
    \end{array}
    \label{eq:score}
  \end{equation}
where $w$ is the set of all path and floorplan waypoints, and where
$a_{ij}$ is the index in $w$ of the $j$th argument of the $i$th atomic formula.
%

%
The transition matrix for the HMM is constructed from the sentences in the
training corpus to allow each state only to self loop or to transition to the
state for the next path prepositional phrase in the training sentence. 
The HMM is constrained to start in the state associated with the first path
prepositional phrase in the sentence associated with each path.
We add dummy states, with a small fixed output probability, between the states
for each pair of adjacent path prepositional phrases, as well as at the
beginning and end of each sentence, to allow for portions of the path that are
not described in the associated sentence.
We then train this HMM with Baum-Welch \citep{Baum1966, Baum1970, Baum1972}.
This trains the distributions for the words in the lexicon~$\Lambda$ as they
are tied as components of the output models.
Specifically, it infers the latent alignment between the noisy robot path
waypoints and the phrases in the training data while simultaneously updating
the meanings of the words to match the relationships between waypoints
described in the corpus.
In this way, the meanings of both the nouns and the prepositions are
learned. 

\subsubsection{Generation}
\begin{figure}
  \centering
  \includegraphics[width=.75\columnwidth]{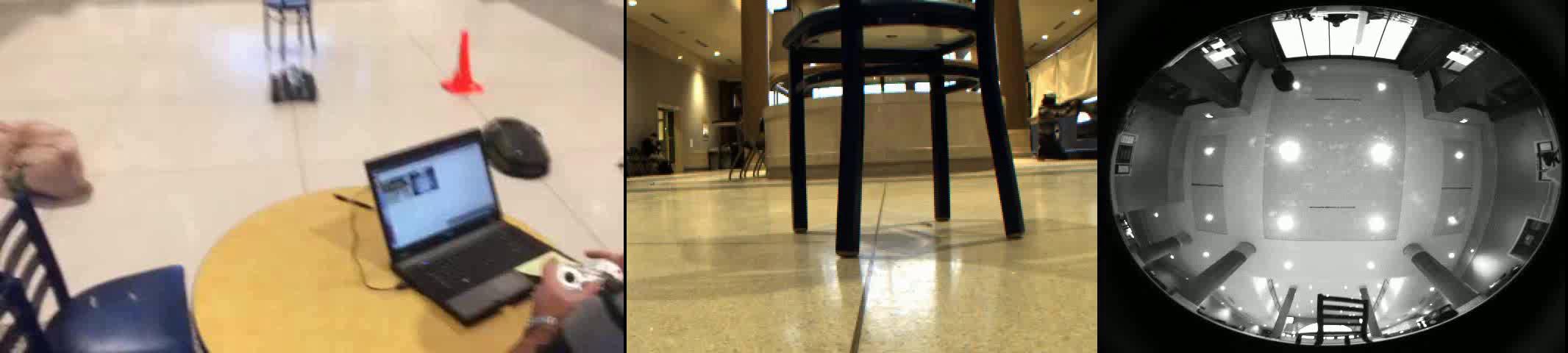}
  \par\vspace*{-2ex}
  \caption{Illustration of the generation algorithm.
    A disambiguating noun phrase is generated for each floorplan waypoint.
    Path waypoints are described by prepositional phrases, and then sets of
    identical phrases are merged into intervals, which are combined to form
    the sentence. 
  }
  \par\vspace*{-2ex}
  \label{fig:generation}
\end{figure}
Language generation takes as input a path~$\mathbf{p}$ obtained by odometry
during human teleoperation of the robot.
This path consists of a collection of 2D floor positions sampled at 50Hz.
To generate a formula in logical form, and thus the corresponding sentence, one
must select a subsequence of this dense sequence worthy of description.

During generation, we care about three properties: ``correctness,'' that the
sentence be logically true of the path, ``completeness,'' that the sentence
differentiate the intended path from all other possible paths, and
``conciseness,'' that the sentence be the shortest that does so.
We attempt to find a balance between these properties with the following
heuristic algorithm (Fig.~\ref{fig:generation}).
First, we sample path waypoints in a way that the sampled points evenly
distribute along the path.
To this end, we downsample the path by computing the integral distance traveled
from the initial position for each point in the dense path and selecting a
subsequence whose points are separated by 5cm of integral path length.
We then produce a path prepositional phrase to describe each path waypoint by
selecting that atomic formula with maximum posterior probability constructed
out of a two-argument predicate with the path waypoint as its first argument
and with a floorplan waypoint as its second argument.
Identical such choices for consecutive sets of waypoints in the path are
coalesced and short intervals of path prepositional phrases are discarded.
We then generate a noun phrase for the object of each waypoint preposition
that refers to that referenced floorplan waypoint.
We take a one-argument predicate to be true of that class with
maximum posterior probability and false of all others.
Similarly, for each pair of floorplan waypoints, we take that two-argument
predicate with maximum posterior probability to be true of that tuple and all
other predicates applied to that tuple to be false.
Thus when the floorplan contains a single instance of a class, it can be
referred to with a simple noun.
But when there are multiple instances of a class, the shortest possible noun
phrase, with one or more SR prepositional phrases, is generated to
disambiguate.

More formally, let~$c(e)$ be the class name of the object at the floorplan
waypoint~$e$.
For each pair of floorplan waypoints~$(e,e_n)$, there exists only one
two-argument spatial-relation predicate $\phi_n$ that is true of this tuple.
%
Let $d(e)$ be the noun phrase we want to generate to disambiguate the floorplan
waypoint~$e$ from others~$e_n$.
Then $e$ can be referred to with $d(e)$ unambiguously if \textbf{(a)} $d(e)=(c(e),\{\})$ is unique; or \textbf{(b)}, there exists a collection of
two-argument predicates $\{\phi_n(e,e_n)\}$ such that formula
$d(e)=\left(c(e),\{(\phi_n,d(e_n))\}\right)$ is unique. 
To produce a concise sentence, we want the size of the collection of
two-argument predicates in step \textbf{(b)} above to be as small as possible.
However, finding the smallest collection of modifiers is NP-hard
\citep{Dale1995}.  
To avoid exhaustive search, we use a greedy heuristic that biases towards
adding the least frequent pairs $(\phi_n,d(e_n))$ into the collection until
$d(e)$ is unique.
This results in a tractable polynomial algorithm.
After we get $d(e)$, we turn it into a noun phrase by simple realization, for
example:
\resizebox{0.5\textwidth}{!}{
  \begin{tabular}{@{}c@{}}
    \textsc{(table, $\{$(left-of, chair), (behind, table)$\}$)}\\
    $\downarrow$\\
    \textit{the table which is left of the chair and behind the table}
  \end{tabular}
}

\subsubsection{Comprehension}

To perform comprehension, we use gradient ascent to optimize the scoring
function with respect to an unknown path~$\mathbf{p}$ 
\begin{equation*}
  \mathbf{p}^*=
  \argmax_{\mathbf{p}}\mathcal{R}(\mathbf{s},\mathbf{p},\mathbf{f},\Lambda)
\end{equation*}
where $\mathcal{R}(\mathbf{s},\mathbf{p},\mathbf{f},\Lambda)$ is the product of
all $R_k$ from Eq.~\ref{eq:score}.
We are computing a MAP estimate of the joint probability of satisfying the
conjunction of atomic formulas assuming that they are independent.

The above scoring function alone is insufficient.
It represents the strict meaning of the sentence, but does not take into account
constraints of the world, such as the need to avoid collision with the objects
in the floorplan.
It can also be difficult to optimize because the cost associated with the
relative orientation between two waypoints becomes increasingly sensitive to
small changes in position as they become closer together.
To remedy the problems of the path waypoints getting too close to objects and to
each other, a barrier penalty term is added between each pair of a path
waypoint and floorplan waypoint as well as between pairs of temporally
adjacent path waypoints to prevent them from becoming too close.   
This term is 1 until the distance between the two waypoints becomes less than a
threshold, at which point it decreases rapidly.
Finally, our formulation of the semantics of prepositions is based on angles
but not distance.
Thus there is is a large subspace of the floor that leads to equal probability
of satisfying each atomic formula, \ie\ the cones in Fig.~\ref{fig:overview}.
This allows a path to satisfy a prepositional phrase like \emph{to the left of
  the chair} by being far away from the chair.
To remedy this, we add a small attraction between each path waypoint and the
floorplan waypoints selected as its reference objects to prefer short
distances.
A postprocessing step performs obstacle avoidance by adding additional path
waypoints as needed.

\section{Experiments}

\begin{figure*}
  \centering
  \resizebox{0.84\textwidth}{!}{\begin{tabular}{@{}rrr@{\hspace*{2pt}}c@{\hspace*{2pt}}c@{\hspace*{2pt}}c@{\hspace*{2pt}}c@{\hspace*{2pt}}c@{\hspace*{2pt}}c@{}}
    \rotatebox{90}{\makebox[0pt]{\hspace*{-180pt}{\textbf{acquisition}}}}&
    \rotatebox{90}{\makebox[0pt]{{\textbf{input}}}}&&
    \includegraphics[height=100pt]{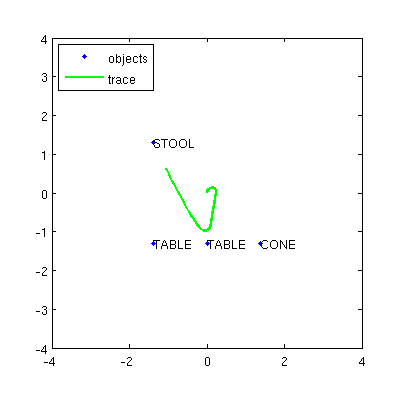}&
    \includegraphics[height=100pt]{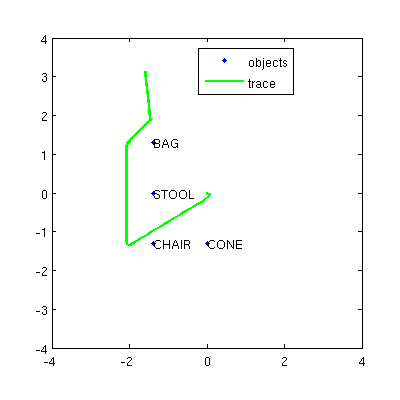}&
    \includegraphics[height=100pt]{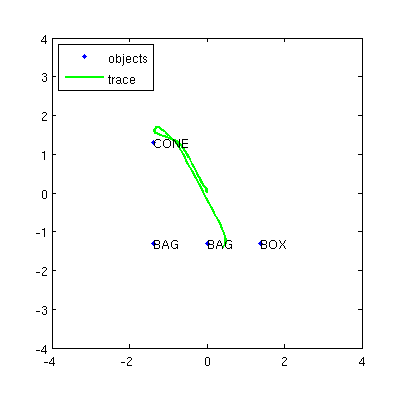}&
    \includegraphics[height=100pt]{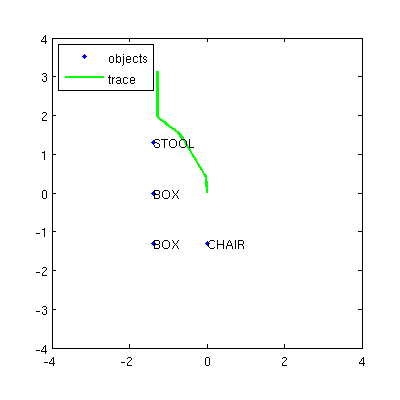}&
    \includegraphics[height=100pt]{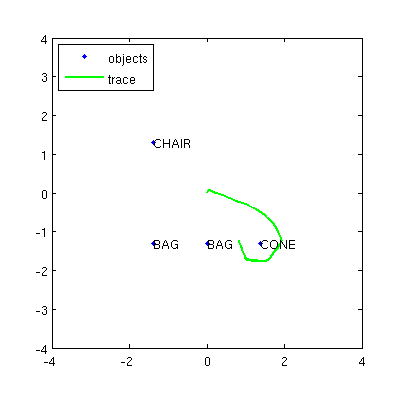}&
    \includegraphics[height=100pt]{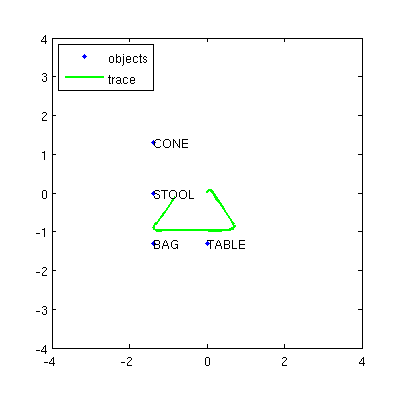}\\
    &&&
    \parbox[t]{0.2\textwidth}{\emph{The robot went towards the table which is
        right of the table then went in front of the stool.}}&
    \parbox[t]{0.2\textwidth}{\emph{The robot went left of the chair then went
        behind the bag then went away from the chair.}}&
    \parbox[t]{0.2\textwidth}{\emph{The robot went behind the cone then went
        right of the bag which is right of the bag.}}&
    \parbox[t]{0.2\textwidth}{\emph{The robot went behind the stool then went
        away from the box which is behind the box.}}&
    \parbox[t]{0.2\textwidth}{\emph{The robot went in front of the cone then
        went right of the bag which is right of the bag.}}&
    \parbox[t]{0.2\textwidth}{\emph{The robot went away from the cone then went
        behind the bag then went right of the stool.}}\\
    \\
    &\rotatebox{90}{\makebox[0pt]{\hspace*{0pt}{\textbf{output}}}}
    &\rotatebox{90}{\makebox[0pt]{\hspace*{50pt}{\textbf{nouns}}}}&
    \includegraphics[width=0.15\textwidth]{bag-distribution}&
    \includegraphics[width=0.15\textwidth]{box-distribution}&
    \includegraphics[width=0.15\textwidth]{chair-distribution}&
    \includegraphics[width=0.15\textwidth]{cone-distribution}&
    \includegraphics[width=0.15\textwidth]{stool-distribution}&
    \includegraphics[width=0.15\textwidth]{table-distribution}\\
    &&&
    \emph{bag} & \emph{box} & \emph{chair} & \emph{cone} & \emph{stool} &
    \emph{table}\\
    \\
    &&\rotatebox{90}{\makebox[0pt]{\hspace*{65pt}{\textbf{position}}}}&
    \includegraphics[width=0.15\textwidth]{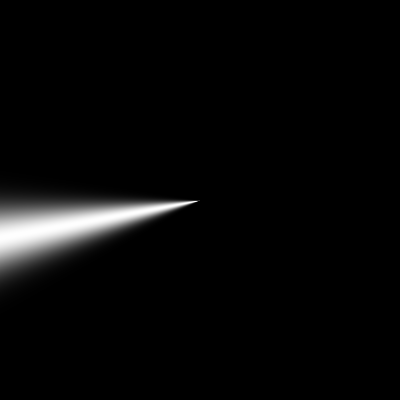}&
    \includegraphics[width=0.15\textwidth]{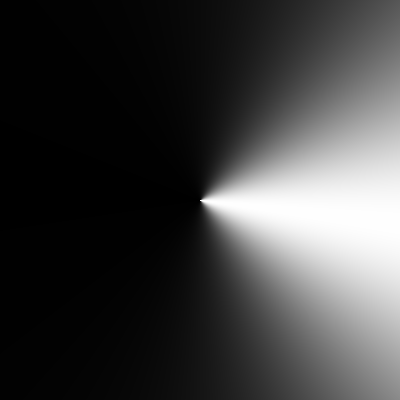}&
    \includegraphics[width=0.15\textwidth]{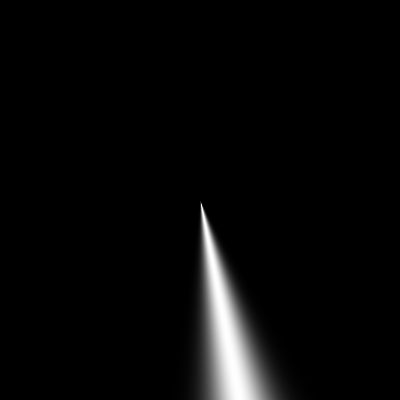}&
    \includegraphics[width=0.15\textwidth]{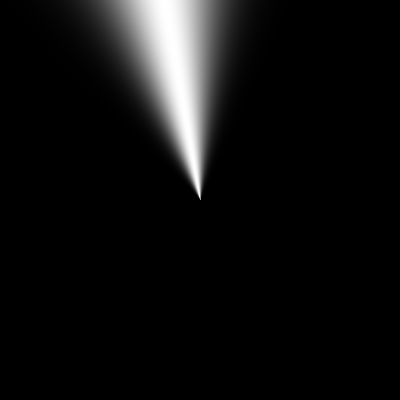}&
    \includegraphics[width=0.15\textwidth]{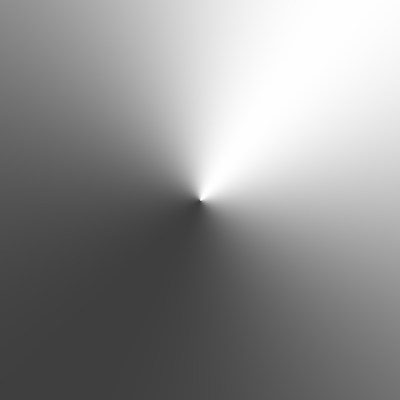}&
    \includegraphics[width=0.15\textwidth]{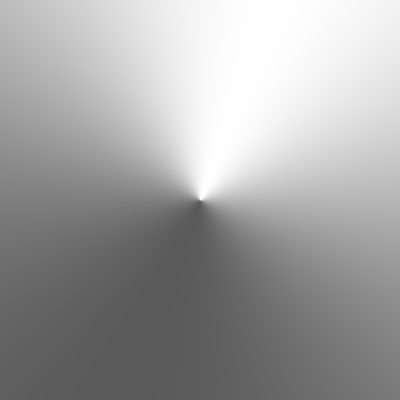}\\
    &&\rotatebox{90}{\makebox[0pt]{\hspace*{65pt}{\textbf{velocity}}}}&
    \includegraphics[width=0.15\textwidth]{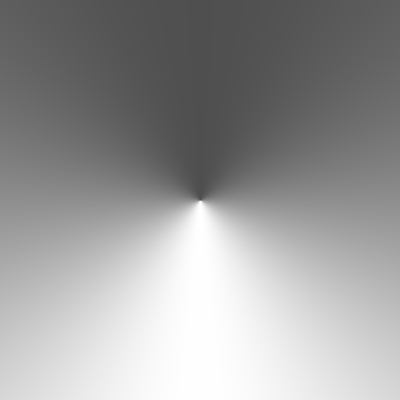}&
    \includegraphics[width=0.15\textwidth]{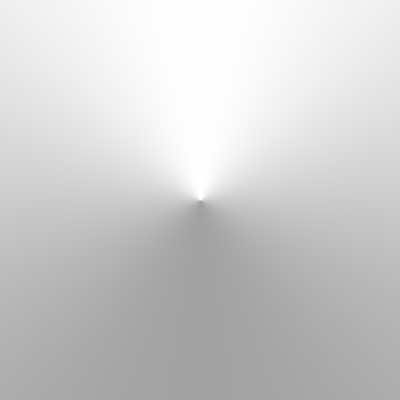}&
    \includegraphics[width=0.15\textwidth]{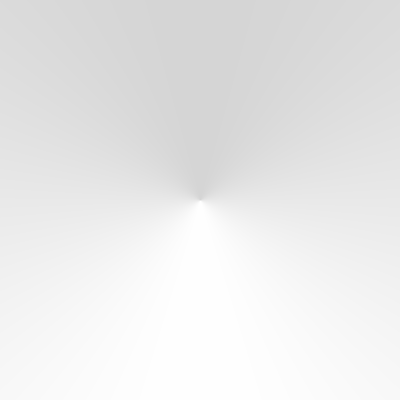}&
    \includegraphics[width=0.15\textwidth]{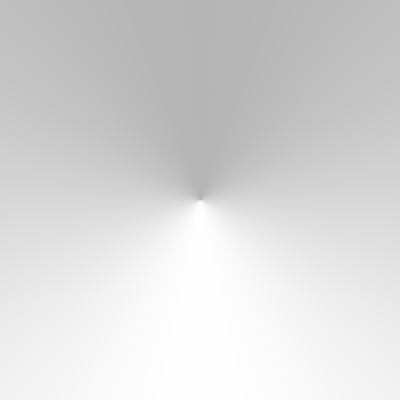}&
    \includegraphics[width=0.15\textwidth]{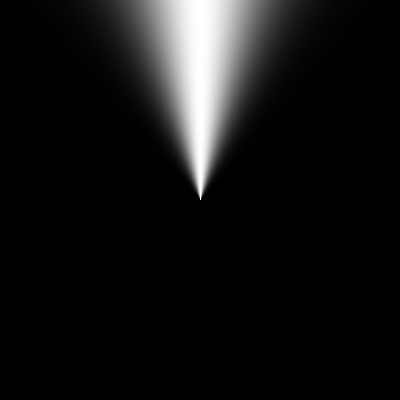}&
    \includegraphics[width=0.15\textwidth]{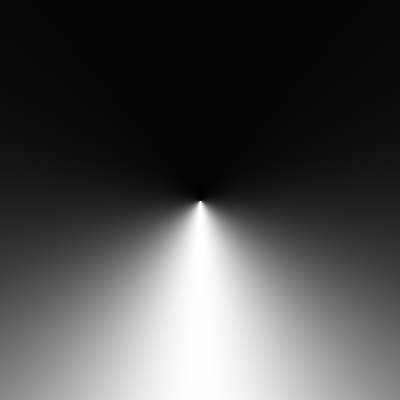}\\
    &&&
    \emph{left of}&
    \emph{right of}&
    \emph{in front of}&
    \emph{behind}&
    \emph{towards}&
    \emph{away from}\\
    \hline\\
    \rotatebox{90}{\makebox[0pt]{\hspace*{-50pt}{\textbf{generation}}}}&
    \rotatebox{90}{\makebox[0pt]{\hspace*{100pt}{\textbf{input}}}}&&
    \includegraphics[height=100pt]{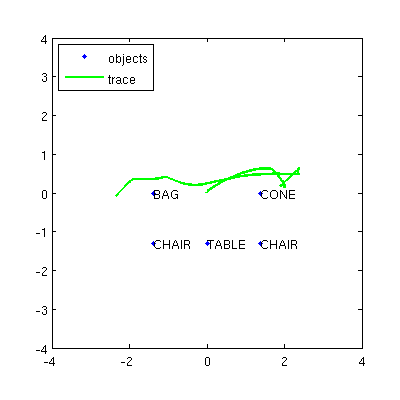}&
    \includegraphics[height=100pt]{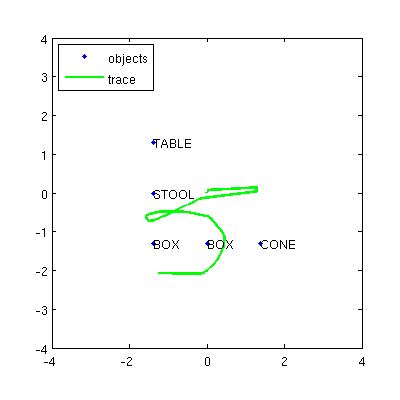}&
    \includegraphics[height=100pt]{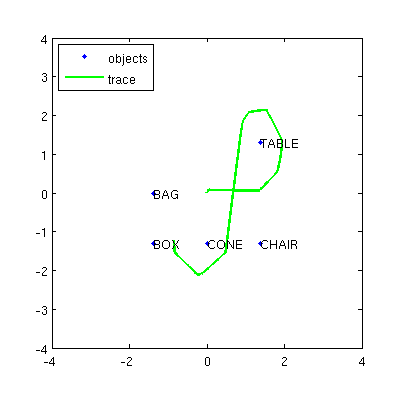}&
    \includegraphics[height=100pt]{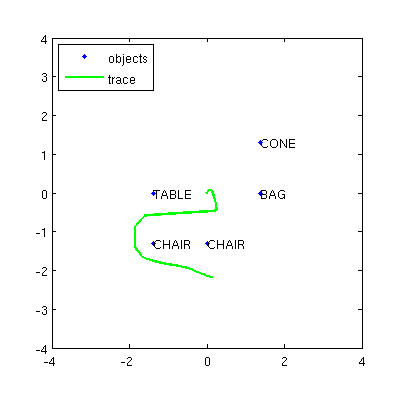}&
    \includegraphics[height=100pt]{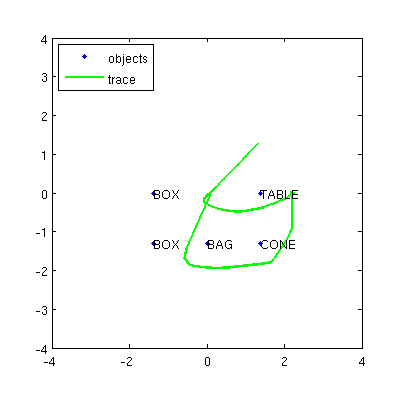}&
    \includegraphics[height=100pt]{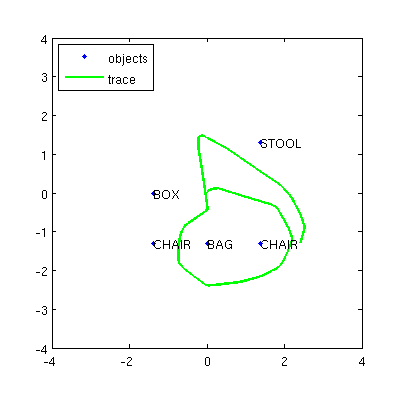}\\
    &\rotatebox{90}{\makebox[0pt]{\hspace*{-100pt}{\textbf{output}}}}&&
    \parbox[t]{0.2\textwidth}{\emph{The robot went behind the cone then went
        away from the cone then went behind the cone then went behind the
        bag.}}&
    \parbox[t]{0.2\textwidth}{\emph{The robot went behind the cone then went in
        front of the stool then went in front of the stool then went right of
        the box which is left of the box then went left of the cone then went
        in front of the box which is right of the box then went in front of the
        box which is left of the box.}}&
    \parbox[t]{0.2\textwidth}{\emph{The robot went in front of the table then
        went right of the table then went behind the table then went left of
        the table then went right of the cone then went in front of the cone
        then went left of the cone.}}&
    \parbox[t]{0.2\textwidth}{\emph{The robot went left of the bag then went
        behind the chair which is right of the chair then went behind the chair
        which is left of the chair then went left of the chair which is left of
        the chair then went in front of the chair which is left of the chair
        then went in front of the chair which is right of the chair.}}&
    \parbox[t]{0.2\textwidth}{\emph{The robot went behind the bag then went
        left of the bag then went in front of the bag then went in front of the
        cone then went behind the cone then went behind the bag then went
        behind the table.}}&
    \parbox[t]{0.2\textwidth}{\emph{The robot went in front of the stool then
        went right of the chair which is right of the bag then went in front of
        the chair which is right of the bag then went in front of the bag then
        went left of the bag then went behind the bag then went away from the
        bag then went left of the stool then went in front of the stool then
        went right of the chair which is right of the bag.}}\\
    \hline\\
    \rotatebox{90}{\makebox[0pt]{\hspace*{-250pt}{\textbf{comprehension}}}}&
    \rotatebox{90}{\makebox[0pt]{\hspace*{-60pt}{\textbf{input}}}}&&
    \parbox[t]{0.2\textwidth}{\emph{The robot went left of the stool then went
        towards the cone then went behind the table which is right of the bag
        then went in front of the stool.}}&
    \parbox[t]{0.2\textwidth}{\emph{the robot went towards the bag then went
        away from the table then went in front of the box then went towards the
        chair.}}&
    \parbox[t]{0.2\textwidth}{\emph{The robot went towards the bag then went
        towards the stool then went towards the table which is left of the
        stool then went in front of the bag.}}&
    \parbox[t]{0.2\textwidth}{\emph{The robot went away from the table which is
        behind the box then went right of the stool then went right of the
        table which is behind the box then went towards the table which is left
        of the box.}}&
    \parbox[t]{0.2\textwidth}{\emph{The robot went towards the bag which is
        left of the stool then went towards the table then went behind the
        table then went left of the bag which is left of the stool.}}&
    \parbox[t]{0.2\textwidth}{\emph{The robot went in front of the chair then
        went in front of the box which is right of the box then went behind the
        box which is right of the box then went towards the box which is left
        of the box.}}\\
    &\rotatebox{90}{\makebox[0pt]{\hspace*{100pt}{\textbf{output}}}}&&
    \includegraphics[height=100pt]{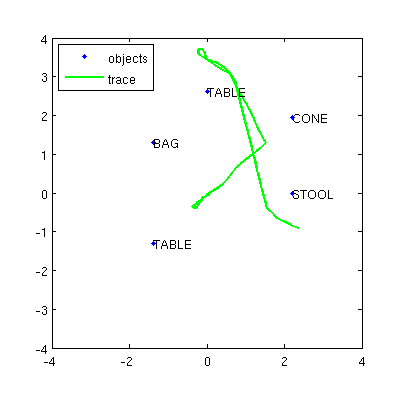}&
    \includegraphics[height=100pt]{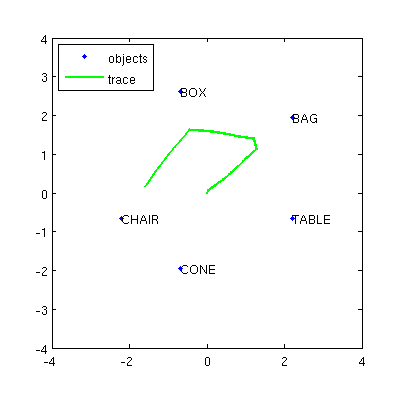}&
    \includegraphics[height=100pt]{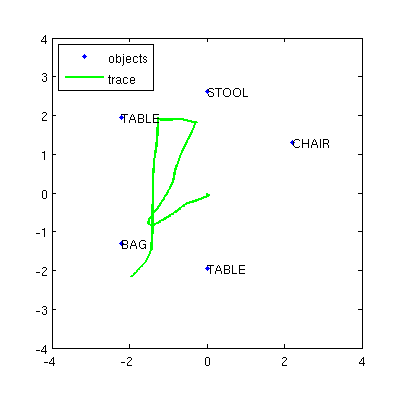}&
    \includegraphics[height=100pt]{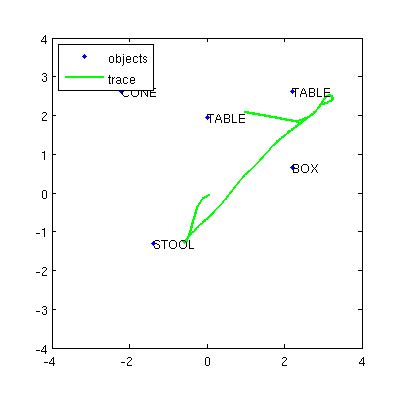}&
    \includegraphics[height=100pt]{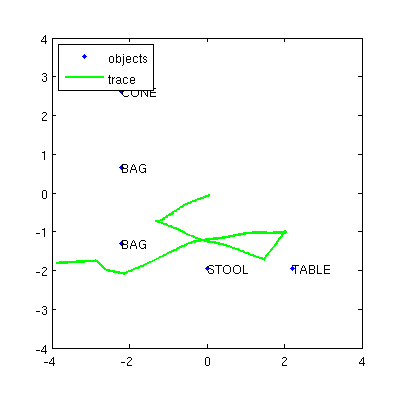}&
    \includegraphics[height=100pt]{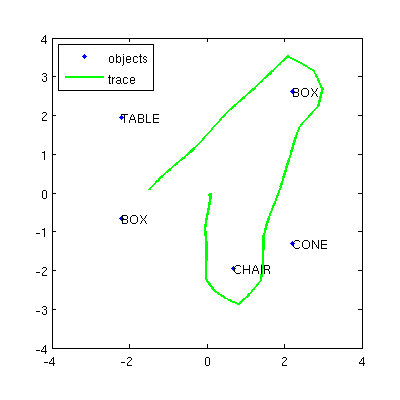}
  \end{tabular}}
  \caption{Example experimental runs, 6 out of 250 for acquisition and 100 for
    each of generation and comprehension.
    Videos available at \url{http://drivingundertheinfluenceoflanguage.blogspot.com}.
    %
  }
  \vspace*{-2ex}
  \label{fig:experiments}
\end{figure*}

We conducted an experiment as outlined in Fig.~\ref{fig:overview}.
We generated 250 random sentences from the grammar in Fig.~\ref{fig:grammar}, 25
in each of 10 different floorplans that were randomly generated to place either
4 or 5 objects, with 2 objects always being of the same class, to introduce
ambiguity requiring disambiguation via SR prepositional phrases, at one of 12
possible grid positions.
Path data was logged while a human teleoperator manually drove the robot to
comply with these sentential instructions in these floorplans
(Fig.~\ref{fig:experiments} top).
Models were learned for each of the nouns and prepositions.
These were used to automatically generate descriptions for 10 different new
paths manually driven by a human teleoperator in 10 new random floorplans
(Fig.~\ref{fig:experiments} middle).
These were also used to automatically drive the robot to follow 10 different
new random sentences in each of 10 different new random floorplans where the
same objects could be placed at one of 56 possible grid positions
(Fig.~\ref{fig:experiments} bottom). 
The random sentences used for training had either 2 or 3 path waypoints while
those used for generation and comprehension had either 5 or 6 path waypoints.

\begin{figure}
  \centering
  \resizebox{.75\columnwidth}{!}{\begin{tabular}{@{}l|rrrr@{}}
    & \multicolumn{2}{c}{correctness}
    & \multicolumn{2}{c}{completeness}\\
    & \multicolumn{1}{c}{mean} & \multicolumn{1}{c}{std dev}
    & \multicolumn{1}{c}{mean} & \multicolumn{1}{c}{std dev}\\
    \hline
    generation (hand-constructed models) & 94.6\% & 4.54\% & 85.5\% & 2.26\%\\
    generation (learned models) & 92.0\% & 6.11\% & 84.2\% & 6.35\%\\
    comprehension (planned path) & 96.2\% & 0.38\% & 88.5\% & 11.5\%\\
    comprehension (measured path) & 95.5\% & 1.42\% & 84.7\% & 9.9\%\\
  \end{tabular}}\\
  \resizebox{.75\columnwidth}{!}{\begin{tabular}{@{}cc@{}}
      \includegraphics{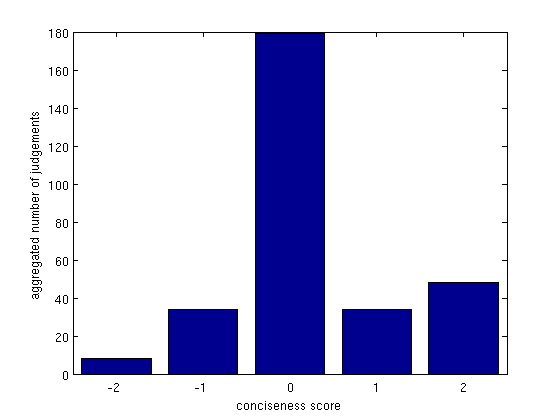}&\includegraphics{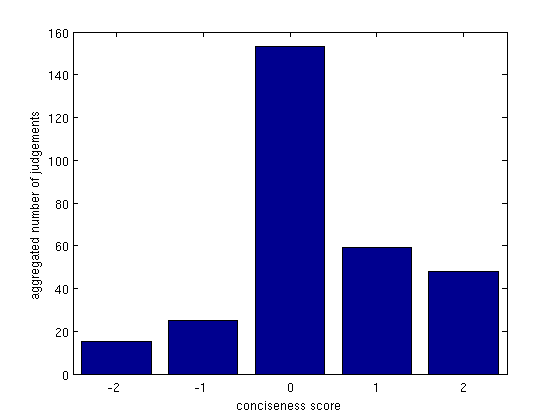}\\
      (hand-constructed models) & (learned models)
    \end{tabular}}
  \par\vspace*{-2ex}
  \caption{Correctness, completeness, and conciseness results of human
    evaluation of sentences automatically generated from manually driven paths
    and automatically driven paths produced by comprehension of provided
    sentences.}
  \par\vspace*{-2ex}
  \label{fig:results}
\end{figure}

Odometry and inertial guidance were used to determine paths driven.
Pairs of sentences and paths obtained during both generation and comprehension
were given to a pool of 6 independent judges to obtain 3 judgments on each.
Judges were asked to label each path prepositional phrase in each sentence
paired with the entire path as being either `correct' or `incorrect',
\ie\ whether it was true of the intended portion of the path as determined by
that judge.
For generation, judges were also asked to assess how much of the path was
described by the sentence, giving a completeness judgment ranging from 0
(worst) to 5 (best).
These were converted to percentages.
For comprehension, judges were also asked to assess what fraction of the path
constitutes motion that is described by the sentence (quantized as 0 to 5).
These were again converted to percentages to measure completeness.
For generation, judgments were obtained twice, pairing each input path with
sentences generated using the hand-constructed models from
Fig.~\ref{fig:overview} as well the learned models from
Fig.~\ref{fig:experiments}.
For comprehension, judgments were also obtained twice, pairing each input
sentence with both the planned path as well as the actually driven path as
determined by odometry and inertial guidance.
Fig.~\ref{fig:results}(top) summarizes the judgments aggregated across the 3
judges and 100 samples.
The standard deviations are across the mean value of the 3 judges for each
sample.
Overall, the average ``correctness'' reported is 94.6\% and the average
``completeness''  reported is 85.6\%.

For generation, we also measured ``conciseness'' by having the 3 human judges
score each generated sentence as -2 (much too short), -1 (too short), 0 (about
right), 1 (too long), or 2 (much too long).
Fig.~\ref{fig:results}(bottom) summarize these judgments as histograms.
Overall, judges assessed that the generated sentence length was `about right' a
little over half of the time, with generation erring more towards being too
long than too short.

\nocite{Branavan2010}
\nocite{carpenter1997}
\nocite{Clarke2010}
\nocite{he2005}
\nocite{hockenmaier2002}
\nocite{Kalman1960}
\nocite{Kollar2010}
\nocite{lecun1998}
\nocite{siddharth2014}
\nocite{steedman1996}
\nocite{steedman2000}

\section{Conclusion}

We demonstrate a novel approach for grounding the semantics of natural
language in the domain of robot navigation.
Sentences describe paths taken by the robot relative to other objects in the
environment.
The meanings of nouns and prepositions are trained from a corpus of paths
driven by a human teleoperator annotated with sentential descriptions.
These can then support both automatic generation of sentential descriptions of
new paths driven as well as automatic driving of paths to satisfy navigational
goals specified in provided sentences.
This is a step towards the ultimate goal of grounded natural language that
allows machines to interact with humans when the language refers to actual
things and activities in the real world.

\iffinal
\section*{Acknowledgments}

This research was sponsored, in part, by the Army Research Laboratory and was
accomplished under Cooperative Agreement Number W911NF-10-2-0060.
The views and conclusions contained in this document are those of the authors
and should not be interpreted as representing the official policies, either
express or implied, of the Army Research Laboratory or the U.S. Government.
The U.S. Government is authorized to reproduce and distribute reprints for
Government purposes, notwithstanding any copyright notation herein.
\fi

\bibliography{acl2015}

\end{document}